\documentclass{article}

\usepackage[preprint]{neurips_2026}

\usepackage[utf8]{inputenc} 
\usepackage[T1]{fontenc}    
\usepackage{hyperref}       
\usepackage{url}            
\usepackage{booktabs}       
\usepackage{amsfonts}       
\usepackage{nicefrac}       
\usepackage{microtype}      
\usepackage{xcolor}         
\usepackage{graphicx}
\usepackage{multirow}       
\usepackage{float}
\usepackage{subcaption}
\usepackage{amsmath}
\usepackage{dblfloatfix}
\usepackage{enumitem}
\usepackage[title]{appendix}
\usepackage{xspace}

\newcommand{\metakv}{\textsc{MetaKV}\xspace}
\newcommand{\static}{\textsc{StaticBaseline}\xspace}
\newcommand{\oracle}{\textsc{Oracle}\xspace}

\title{\metakv: Adaptive KV Cache Compression for Constrained LLM Inference}

\author{
  Michael Wang\thanks{These authors contributed equally.} \\
  Lake Washington School District \\
  \texttt{1104989@lwsd.org}
  \And
  Keith Li\footnotemark[1] \\
  XSchool \\
  \texttt{keith.li@x-school.org}
  \And
  Roozbeh Bostandoost\thanks{Corresponding author: \texttt{rbostandoost@umass.edu}.} \\
  UMass Amherst \\
  \texttt{rbostandoost@umass.edu}
}

\begin{document}

\maketitle

\begin{abstract}

    Key--value (KV) cache compression is an effective way to reduce the memory overhead of large language model (LLM) inference, particularly for long-context workloads. However, existing compression methods make different trade-offs among accuracy, inference latency, and peak KV cache memory utilization, making a single fixed configuration unsuitable across different prompts and resource constraints. We introduce \metakv, an adaptive framework that selects a KV cache compression configuration for each input prompt based on user-specified latency and peak memory budgets. \metakv uses lightweight prediction models to estimate the end-to-end latency, peak memory, and probability of a correct response for each candidate configuration, and selects the configuration that best satisfies the latency-memory constraints while preserving accuracy. We evaluate \metakv across ten configurations from three representative KV cache compression methods, KVQuant, H$_2$O, and RocketKV, together with an uncompressed FP16 configuration, on four datasets covering mathematics, science, commonsense reasoning, and reading comprehension. Across a wide range of latency and peak memory constraints, \metakv consistently outperforms the best static configuration, improving constrained success rate (CSR), the fraction of prompts answered correctly while satisfying both constraints, by approximately 0.07 on average and up to 0.135. These results demonstrate the benefit of adapting KV cache compression to individual prompts and latency-memory constraints. Code is available at \url{https://github.com/MichaelWang0505/MetaKV.git}.
\end{abstract}

\section{Introduction}


Large Language Models (LLMs) have become widely used for applications such as conversational assistants and code generation~\cite{Zhao_2026, dong2025surveycodegenerationllmbased}. Their growing adoption has increased the demand for GPU memory and efficient LLM inference~\cite{Oviedo_2026, patel2024splitwiseefficientgenerativellm, agrawal2023sarathiefficientllminference}. This challenge is especially important for long-context inference, where the KV cache, which stores the key and value vectors of previously processed tokens, can consume up to $80\%$ of GPU memory~\cite{hooper2025kvquant10millioncontext}. To reduce this memory overhead and serve more requests on limited GPU resources, prior work has proposed KV cache compression methods. These methods can be broadly categorized as lossless or lossy. Lossless methods reduce memory usage while preserving the original KV cache values and, therefore, model accuracy. Lossy methods reduce precision or discard parts of the KV cache, achieving larger memory savings at the cost of some accuracy loss. Since long-context inference often requires substantial memory reduction and lossless compression provides limited savings, we focus on lossy KV cache compression in this work.

State-of-the-art lossy KV cache compression methods use different approaches. These include (i) \textit{quantization}, reducing the bits used to store KV values~\cite{hooper2025kvquant10millioncontext, https://doi.org/10.13140/rg.2.2.28167.37282, he2024zipcacheaccurateefficientkv}; (ii) \textit{token eviction}, permanently removing less important tokens from the KV cache~\cite{zhang2023H2Oheavyhitteroracleefficient}; and (iii) \textit{sparse attention}, attending to only a subset of cached tokens during computation~\cite{behnam2025rocketkvacceleratinglongcontextllm, li2024snapkvllmknowslooking}. Each approach makes different trade-offs among peak KV cache memory utilization (peak memory), end-to-end latency, and accuracy.

Quantization can substantially reduce peak memory by representing values with only a few bits instead of full 16-bit floating-point precision (FP16). But the additional quantization and dequantization operations can increase inference latency. In contrast, token eviction and sparse attention reduce computation by removing or skipping tokens, often resulting in lower latency. However, the retained KV values typically remain at higher precision, leading to larger peak memory usage than aggressive quantization. So, no single compression approach is consistently superior across all efficiency and accuracy objectives.

Different LLM deployment environments also impose different requirements on accuracy, end-to-end latency, and peak memory. For example, LLMs are increasingly deployed on edge devices such as smartphones, where limited memory makes peak memory usage a major constraint~\cite{xue2024powerinfer2fastlargelanguage}. In contrast, interactive applications such as customer-service chatbots require fast responses, making inference latency a key constraint~\cite{gnewuch2022response}. These scenarios can favor different KV cache compression methods. Quantization may be more suitable for memory-constrained environments, while token eviction may better suit latency-constrained settings. Sparse attention can also be effective because it has the least latency on shorter-context prompts~\cite{behnam2025rocketkvacceleratinglongcontextllm}. Compression methods can also perform differently across datasets because the datasets emphasize different capabilities. Since different compression methods preserve or remove information from the KV cache in different ways, their relative performance can vary across these task types. For example, token eviction can achieve higher accuracy than quantization on math prompts, while quantization can perform better on scientific-reasoning prompts. This difference can occur because token eviction removes entire token representations, while quantization retains all tokens at reduced precision, causing the two approaches to preserve task-relevant information differently. Per-dataset results can be viewed in \autoref{app:additional-results}.

Since the relative performance of compression methods can vary across prompt types and deployment constraints, we propose \metakv, an adaptive framework that selects the most suitable KV cache compression configuration for each input prompt.
Given end-to-end latency and peak memory constraints from the user, \metakv predicts the accuracy, end-to-end latency, and peak KV cache memory of each candidate configuration and selects the one that best satisfies the constraints while preserving accuracy. This allows \metakv to adapt its compression strategy to both the input prompt and the deployment environment.

In summary, this paper makes three main contributions. 
First, we show that existing KV cache compression methods exhibit different trade-offs across accuracy, end-to-end latency, and peak memory utilization, motivating the need for adaptive selection. Second, we introduce \metakv, a framework that predicts the accuracy, latency, and peak KV cache memory of candidate compression configurations and selects the most suitable one under given constraints. 
Third, we evaluate \metakv across multiple KV cache compression methods and datasets, showing that it consistently outperforms the best static configuration across all tested latency and peak memory constraint pairs, improving constrained success rate, the fraction of prompts answered correctly while satisfying both constraints, by approximately 0.07 on average and up to 0.135.

\section{Designing \metakv}
\label{sec:metaKV}
In this section, we first review the KV cache compression methods considered in our study. We then compare their performance and show that they make different trade-offs among accuracy, latency, and peak KV cache memory, with no single method performing best across all settings. Motivated by this observation, we present \metakv, which selects the most suitable compression configuration for each prompt under given resource constraints.

\subsection{Baseline KV cache compression methods}
We consider KVQuant~\cite{hooper2025kvquant10millioncontext}, H$_2$O~\cite{zhang2023H2Oheavyhitteroracleefficient}, and RocketKV~\cite{behnam2025rocketkvacceleratinglongcontextllm} because they represent three major categories of lossy KV cache compression: quantization, token eviction, and sparse attention. These methods provide \metakv with a diverse set of trade-offs among accuracy, latency, and peak memory usage.

\noindent\textbf{KVQuant~\cite{hooper2025kvquant10millioncontext}:}
KVQuant reduces KV cache memory by storing Key and Value vectors with fewer bits instead of full 16-bit precision (FP16). Lower precision can reduce model accuracy, so KVQuant includes techniques to limit this loss. For example, it quantizes Keys per channel based on the observation that outliers are concentrated in consistent channels, and stores outlier values in uncompressed FP16 to avoid large quantization errors. We denote a KVQuant setting as $b$-bit and evaluate $b \in \{2, 3, 4\}$, where $b$ is the number of bits used to store KV cache values.

\noindent\textbf{H$_2$O~\cite{zhang2023H2Oheavyhitteroracleefficient}:}
Instead of reducing precision, H$_2$O removes the Key and Value vectors of less important tokens. It tracks an accumulated attention score for each token, reflecting how much attention the token receives from subsequent tokens. When the KV cache exceeds a given budget, H$_2$O evicts older tokens with the lowest accumulated attention scores. We denote an H$_2$O setting as $p\%$ and evaluate $p \in \{20, 40, 60\}$, where $p$ is the percentage of tokens retained in the KV cache relative to the tokens in the input prompt.

\noindent\textbf{RocketKV~\cite{behnam2025rocketkvacceleratinglongcontextllm}:}
RocketKV combines token pruning with sparse attention. In its first stage, SnapKV uses aggregated attention scores between the input context and an observation window at the end of the prompt to identify and retain the most relevant tokens. In its second stage, Hybrid Sparse Attention attends to only a subset of earlier tokens during each decoding step. We denote a RocketKV setting as $r\times$ and evaluate $r \in \{8, 16, 32\}$, where $r$ is the post-prefill compression ratio, meaning that the effective KV cache is reduced to approximately $1/r$ of its original size after prefill.

\subsection{Motivating \metakv}

To compare the three compression methods with the uncompressed baseline, we measure three key metrics for each configuration: accuracy, end-to-end latency, and peak KV cache memory. These results reveal clear trade-offs across methods and motivate the need for \metakv.

\begin{table}[t]
  \centering
  \caption{Accuracy, average end-to-end latency, and average peak memory for
  each configuration over the 1,232 held-out test prompts. The uncompressed
  configuration uses the full KV cache in FP16. Setting denotes the
  quantization bit-width for KVQuant, fraction of tokens retained for H$_2$O,
  and post-prefill compression ratio for RocketKV.}
  \label{tab:tradeoffs}
  \begin{tabular}{llccc}
    \toprule
    Method & Setting & Accuracy & Latency (s) & Peak Memory (MB) \\
    \midrule
    Uncompressed & -- & 0.645 & 2.01 & 34.47 \\
    \midrule
    \multirow{3}{*}{KVQuant}
      & 4-bit & 0.630 & 21.67 & 11.71 \\
      & 3-bit & 0.609 & 20.74 &  8.91 \\
      & 2-bit & 0.536 & 19.68 &  6.39 \\
    \midrule
    \multirow{3}{*}{H$_2$O}
      & 60\% & 0.627 & 2.48 & 19.09 \\
      & 40\% & 0.602 & 2.58 & 12.70 \\
      & 20\% & 0.507 & 2.70 &  6.29 \\
    \midrule
    \multirow{3}{*}{RocketKV}
      & $8\times$  & 0.464 & 3.56 & 30.77 \\
      & $16\times$ & 0.352 & 3.53 & 30.65 \\
      & $32\times$ & 0.283 & 3.83 & 30.64 \\
    \bottomrule
  \end{tabular}
\end{table}

\autoref{tab:tradeoffs} summarizes these trade-offs over 8,192 prompts across four datasets, with experimental details provided in \autoref{eval:exp-setup}. Across the tested configurations, KVQuant has approximately $8.0\times$ the latency of H$_2$O and $5.7\times$ the latency of RocketKV, while using about $29\%$ less peak KV cache memory than H$_2$O and $71\%$ less than RocketKV on average. 

These results show a clear trade-off between end-to-end latency, peak memory usage, and accuracy. KVQuant is more suitable for memory-constrained settings because it achieves the lowest peak KV cache memory, but it incurs substantially higher latency. H$_2$O provides much lower latency while still reducing memory usage, making it more suitable when latency is the main constraint. When neither latency nor peak memory is constrained and accuracy is the primary objective, the uncompressed FP16 configuration performs best. On average, RocketKV did not outperform the other methods. However, these averages hide an important point: on individual datasets, the fastest method changes. For example, on multiple-choice scientific reasoning, RocketKV is the fastest, although FP16 is fastest on average. Accuracy can shift by task as well: under aggressive compression, H$_2$O keeps more accuracy than KVQuant on math problems, while KVQuant is more accurate on multiple-choice scientific reasoning. As mentioned in the introduction, compression methods can perform differently across datasets, since they preserve or remove information from the KV cache in different ways, meaning that their performance can vary based on the task or dataset. Performance can also vary across individual prompts because of differences in complexity and context length, so \metakv adapts to different constraints and prompts by choosing different method configurations.

\subsection{\metakv}

\begin{figure}[t]
    \centering
    \includegraphics[width=0.8\textwidth]{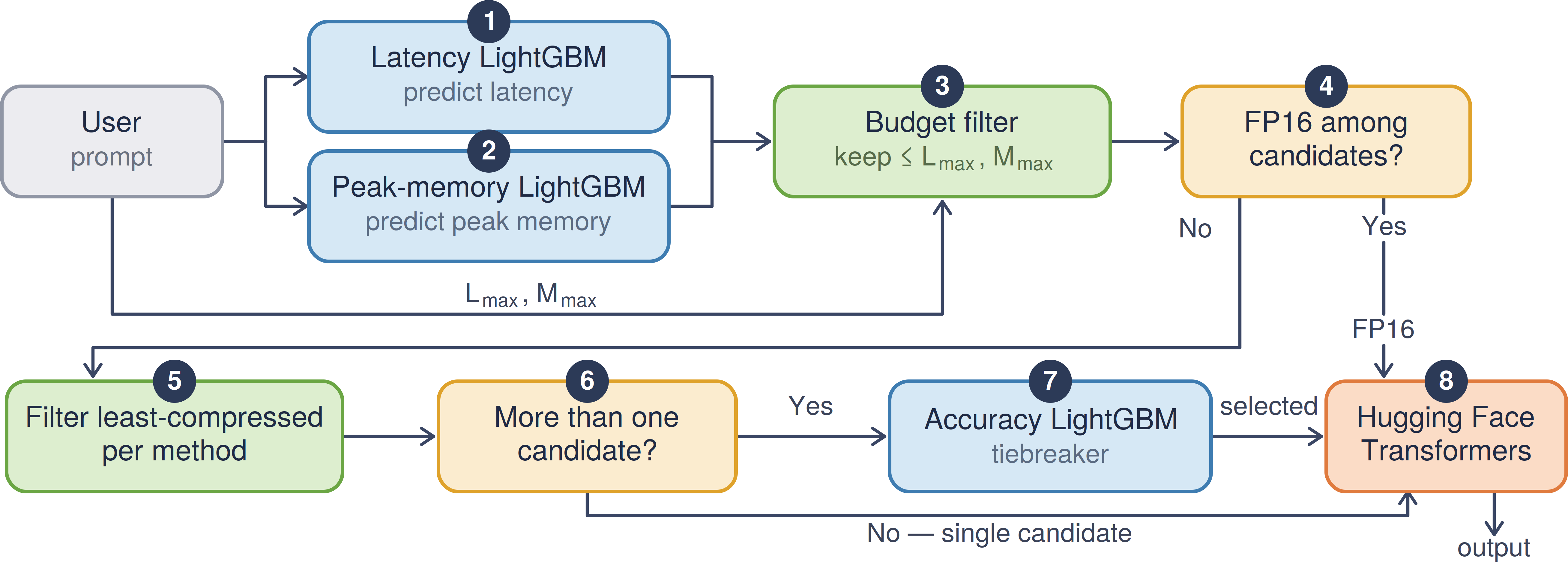}
    \caption{
            The \textsc{MetaKV} pipeline. Given a user's prompt and latency and peak
            memory budgets, the latency and peak memory LightGBMs predict each
            configuration's resource usage, and only those within
            budget are kept. The accuracy LightGBM then predicts which surviving candidate is most
            likely to answer correctly. The selected configuration is executed with Hugging
            Face Transformers.}
    \label{fig:metakv-diagram}
\end{figure}

\metakv selects the best KV cache compression configuration for a given prompt. Its goal is to satisfy user-defined limits for maximum end-to-end latency ($L_{\max}$) and peak memory utilization ($M_{\max}$) while maximizing the likelihood of a correct response. As shown in \autoref{fig:metakv-diagram}, the system follows a straightforward pipeline. First (Steps 1 and 2), when a user submits a prompt alongside their constraints, \metakv uses Light Gradient-Boosting Machine (LightGBM)~\cite{lightgbm} to predict the execution latency and peak memory utilization for all candidate configurations listed in \autoref{tab:tradeoffs}. Next, a Budget Filter evaluates these predictions and discards any configurations that exceed the user's limits (Step 3). 

\metakv then selects the final configuration using a logical hierarchy (Steps 4--7). If the uncompressed FP16 baseline survives the filter, it is selected immediately because it avoids information loss from KV cache compression and achieves the highest overall accuracy. If FP16 is too expensive, \metakv narrows the remaining candidates by keeping only the least compressed setting for each method to preserve as much information as possible. If only one candidate is left, it is selected. If multiple remain, an Accuracy LightGBM predicts their correctness probabilities, and the highest scorer is chosen. Finally, the selected configuration and the user's prompt are sent to the LLM inference engine (Hugging Face Transformers)\cite{wolf2020huggingfacestransformersstateoftheartnatural} to generate the response (Step 8).

If the user's limits are so strict that no configuration passes the Budget Filter, \metakv calculates a relative violation score ($v_c$) to identify which configuration exceeds the limits the least: $v_c = \nicefrac{\hat{L}_c}{L_{\max}} + \nicefrac{\hat{M}_c}{M_{\max}}$, where $\hat{L}_c$ and $\hat{M}_c$ are the predicted end-to-end latency and peak memory utilization for configuration $c$. \metakv then selects the configuration with the smallest combined violation, using the accuracy model to break any ties.

\section{Evaluation}

In this section, we evaluate \metakv's performance and examine how its configuration choices change under different latency and peak memory constraints. We plan to open-source our tool.

\subsection{Experimental setup}
\label{eval:exp-setup}

\noindent\textbf{Model, Hardware, and Datasets.} We evaluated \metakv using Meta Llama 3.1 8B~\cite{grattafiori2024llama3herdmodels} on a single NVIDIA RTX PRO 6000 Blackwell Server Edition GPU (96 GB VRAM) to ensure fair comparisons. To cover a diverse range of workloads, we used four datasets: GSM8K~\cite{cobbe2021trainingverifierssolvemath} (math), HellaSwag~\cite{zellers2019hellaswagmachinereallyfinish} (commonsense), ARC-Challenge~\cite{clark2018thinksolvedquestionanswering} (scientific reasoning), and SQuAD 1.1~\cite{rajpurkar2016squad100000questionsmachine} (reading comprehension). For generative tasks, accuracy was measured using the exact match; for multiple-choice tasks, we selected the answer with the lowest negative log-likelihood and measured whether the selected answer was correct.

\noindent\textbf{Data Collection and Model Training.} For each dataset, we collected 2,048 prompts and profiled ten candidate configurations: uncompressed FP16; KVQuant at 2-, 3-, and 4-bit precision; H$_2$O with 20\%, 40\%, and 60\% token budgets; and RocketKV at 8$\times$, 16$\times$, and 32$\times$ compression ratios. The three settings we chose for each method span mild, moderate, and aggressive compression. This provides clear resource-to-accuracy trade-offs while ensuring the number of configurations is balanced across methods.
We randomly split this data into 85\% training and 15\% testing, yielding 308 test prompts per dataset for a total of 1,232 evaluation prompts. \metakv's LightGBM predictors for latency, peak memory, and accuracy were tuned on the training split using 5-fold cross-validation. The peak memory model achieved high accuracy (MAPE 7.7\% and $R^2$ 0.9981), the latency model performed well overall despite natural variance in RocketKV's dynamic token selection (MAPE 38.19\% and $R^2$ 0.9651, and the accuracy model demonstrated moderate predictive power (pooled AUROC of 0.70). Although running the additional LightGBM predictors before LLM inference
introduces extra latency, their average overhead is only 10.96 miliseconds per prompt, equivalent to 0.34\% of MetaKV's average LLM inference latency, and is therefore negligible.


\noindent\textbf{Evaluation Metrics.} We tracked accuracy, end-to-end latency, and peak KV cache memory. To capture \metakv's ability to balance performance and efficiency, we introduce the Constrained Success Rate (CSR). CSR measures the proportion of test prompts answered correctly without exceeding the user's constraints:
\(
\mathrm{CSR}
=
\frac{1}{N}
\sum_{i=1}^{N}
\mathbb{I}
\left(
\mathrm{correct}_i
\land \ell_i \leq L_{\max}
\land m_i \leq M_{\max}
\right)
\)
where for $N$ test prompts, the indicator function checks if the response is correct ($\mathrm{correct}_i$) and whether the actual latency ($\ell_i$) and peak memory ($m_i$) do not exceed their respective budgets ($L_{\max}$ and $M_{\max}$).


\noindent\textbf{Baselines.} We compared \metakv against two baselines: (i) \static
represents the single best fixed configuration a user could commit to in advance; for any given $(L_{\max},M_{\max})$ constraint pair, we selected the configuration with the highest CSR on the training set and evaluated it on the test set; (ii) \oracle
represents \metakv's theoretical upper bound under perfect predictions; it scores a success if \textit{any} configuration that physically satisfies the true latency and peak memory constraints produces the correct answer.

\subsection{Evaluating \metakv}
\label{eval:results}

\noindent \paragraph{\metakv consistently outperforms \static.} 
As shown in \autoref{tab:tradeoffs}, static compression methods struggle to balance accuracy, end-to-end latency, and peak memory utilization. To quantify the value of adaptive selection, we measured \metakv's Constrained Success Rate (CSR) improvement over the \static ($\Delta\text{CSR} = \text{CSR}_{\metakv} - \text{CSR}_{\static}$) across a grid of latency ($L_{\max}$) and peak memory ($M_{\max}$) limits. 

\begin{figure*}[t]
    \centering

    \begin{minipage}[t]{0.57\textwidth}
        \vspace{0pt}
        \centering
        \includegraphics[width=\linewidth]{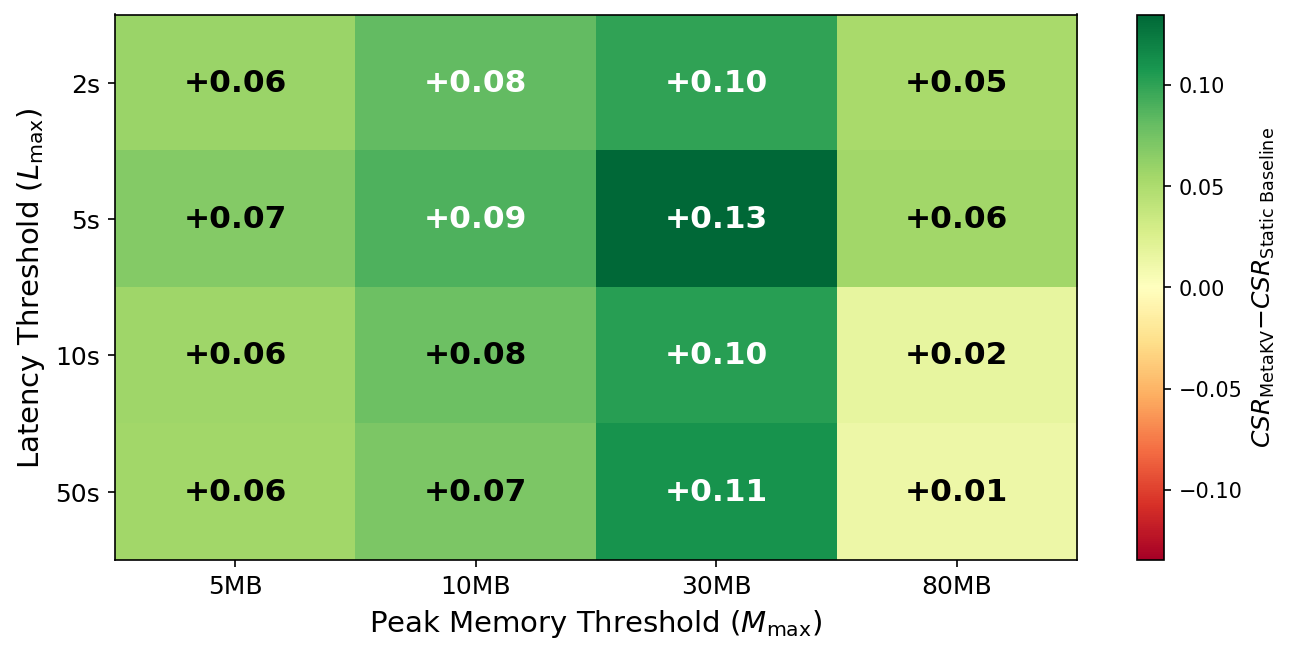}
        \captionof{figure}{\metakv's constrained success rate (CSR) improvement over
                            \static across latency and peak memory constraints, shown as proportions:
                            $\Delta\mathrm{CSR} = \mathrm{CSR}_{\metakv} - \mathrm{CSR}_{\static}$. Positive values
                            indicate \metakv answers more prompts correctly within budget.}
        \label{fig:constraint-heatmap}
    \end{minipage}
    \hfill
    \begin{minipage}[t]{0.40\textwidth}
        \vspace{0pt}
        \centering
        \captionsetup{type=table}

        \captionof{table}{The static configuration with the highest training-set CSR for each
                        latency and peak memory constraint pair. This is the configuration
                        \static commits to for that constraint.}
        \label{tab:constraint-table}

        \renewcommand{\arraystretch}{1.45}
        \setlength{\tabcolsep}{3pt}

        \resizebox{\linewidth}{!}{%
            \begin{tabular}{lcccc}
                \toprule
                & \multicolumn{4}{c}{Peak memory budget} \\
                \cmidrule(lr){2-5}
                Latency & 5 MB & 10 MB & 30 MB & 80 MB \\
                \midrule
                2 s  & Uncompressed & H$_2$O 20\% & H$_2$O 20\% & H$_2$O 60\% \\
                5 s  & H$_2$O 20\% & H$_2$O 40\% & H$_2$O 20\% & H$_2$O 60\% \\
                10 s & H$_2$O 20\% & H$_2$O 40\% & H$_2$O 20\% & H$_2$O 60\% \\
                50 s & H$_2$O 20\% & H$_2$O 40\% & H$_2$O 20\% & H$_2$O 60\% \\
                \bottomrule
            \end{tabular}%
        }

    \end{minipage}

\end{figure*}

\autoref{fig:constraint-heatmap} shows that \metakv achieves positive $\Delta \mathrm{CSR}$ across all tested constraint pairs, with an average improvement of approximately 0.07. 
The largest and most consistent gains occur at the 30~MB peak memory threshold, where \metakv improves CSR by 0.10--0.13 across all latency constraints.
Under the tightest peak memory constraint of 5~MB, \metakv improves CSR by approximately 0.06 on average, while under the tightest latency constraint of 2~s, it improves CSR by approximately 0.073.

\autoref{tab:constraint-table}
shows that the best static configuration under tighter peak memory budgets of 5--30~MB is typically H$_2$O 20\% or 40\%. Across these tighter peak memory budgets, $\Delta\mathrm{CSR}$ increases as the peak memory budget grows from 5~MB to 30~MB. Because the static policy must use the same configuration for every prompt, it cannot take advantage of prompts that could satisfy the constraints with a less aggressive and more accurate configuration. \metakv instead adapts its choice per prompt, allowing it to use lighter compression when the prompt requires fewer resources while using stronger compression when necessary. Once the peak memory budget reaches 80~MB, the \static can finally afford H$_2$O 60\%. This increases the \static's accuracy, narrowing the performance gap ($\Delta\mathrm{CSR}$ drops to $\sim$0.01).



Along the latency axis, \metakv's gains depend on the peak memory budget. At 5--30~MB, tightening the latency threshold does not produce a consistent trend because both \metakv and the \static are restricted to similar low peak memory configurations. At 80~MB, however, \metakv's gain increases from 0.01 at 50~s to 0.052 at 2~s, since the larger peak memory budget gives \metakv more flexibility to select faster configurations for latency-sensitive prompts while preserving accuracy on others.


\noindent \textit{\textbf{Key Takeaways:} \metakv improves CSR across every tested latency and peak memory constraint, with the largest gains when resource constraints create meaningful trade-offs among configurations. Its per-prompt selection allows it to use stronger compression when needed and preserve more accuracy when resources allow.}

\begin{figure*}[t]
    \centering

    \begin{subfigure}[t]{0.47\textwidth}
        \centering
        \phantomsubcaption
        \label{fig:corner-csr}

        \includegraphics[width=\linewidth]
        {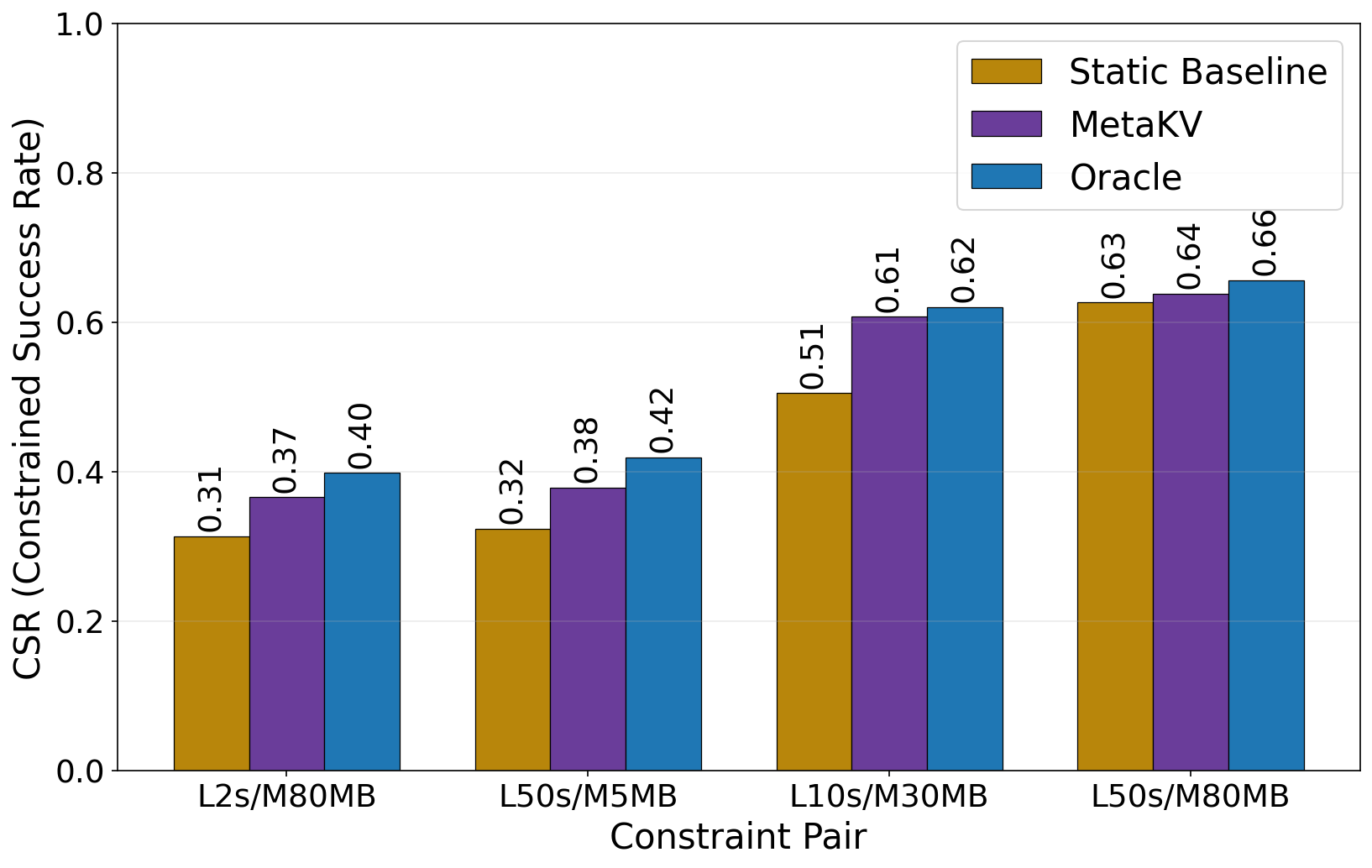}

        \textbf{(a)}
    \end{subfigure}
    \hfill
    \begin{subfigure}[t]{0.47\textwidth}
        \centering
        \phantomsubcaption
        \label{fig:corner-violations}

        \includegraphics[width=\linewidth]
        {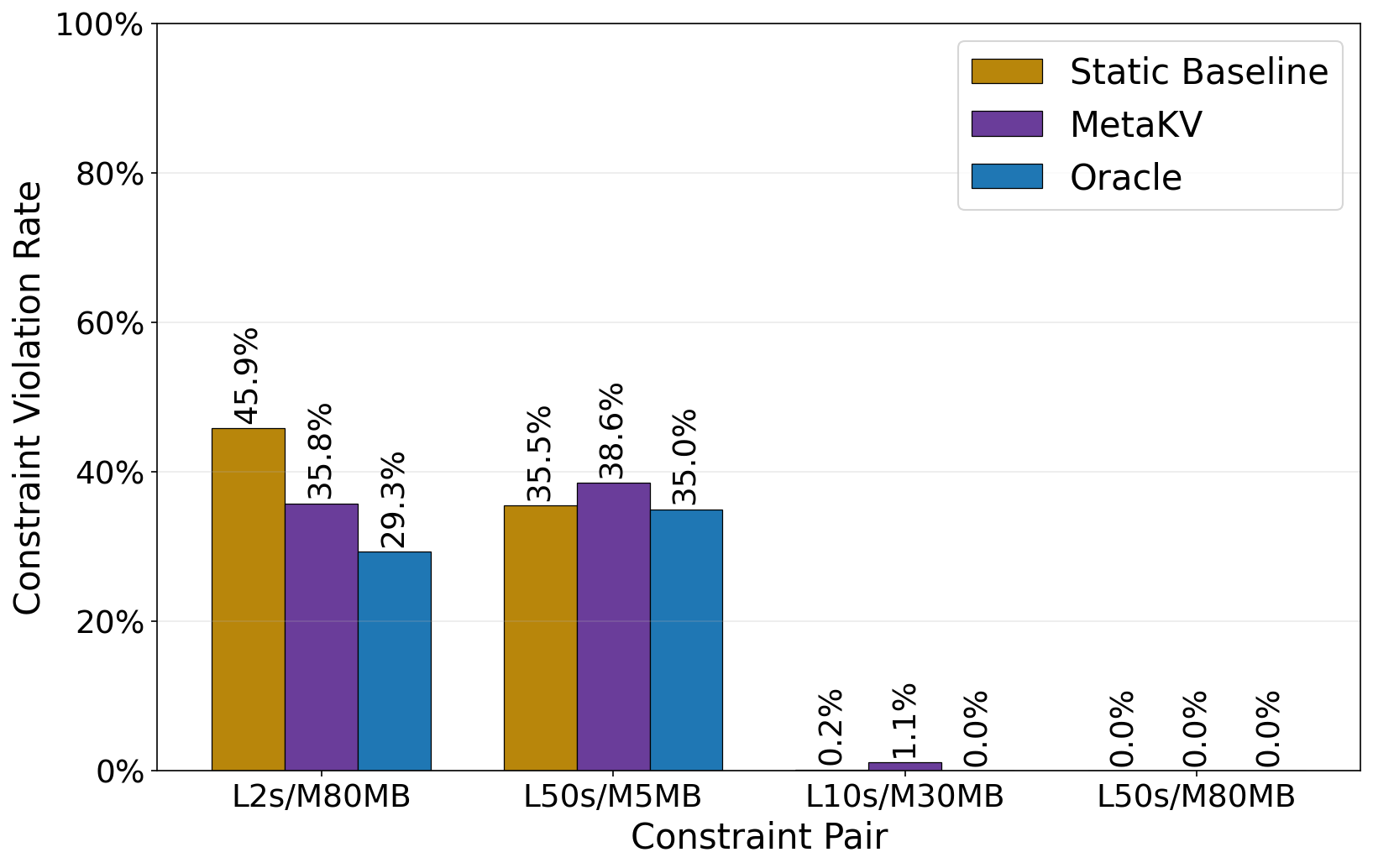}

        \textbf{(b)}
    \end{subfigure}
    \hfill
    \begin{subfigure}[t]{0.47\textwidth}
        \centering
        \phantomsubcaption
        \label{fig:corner-accuracy}

        \includegraphics[width=\linewidth]
        {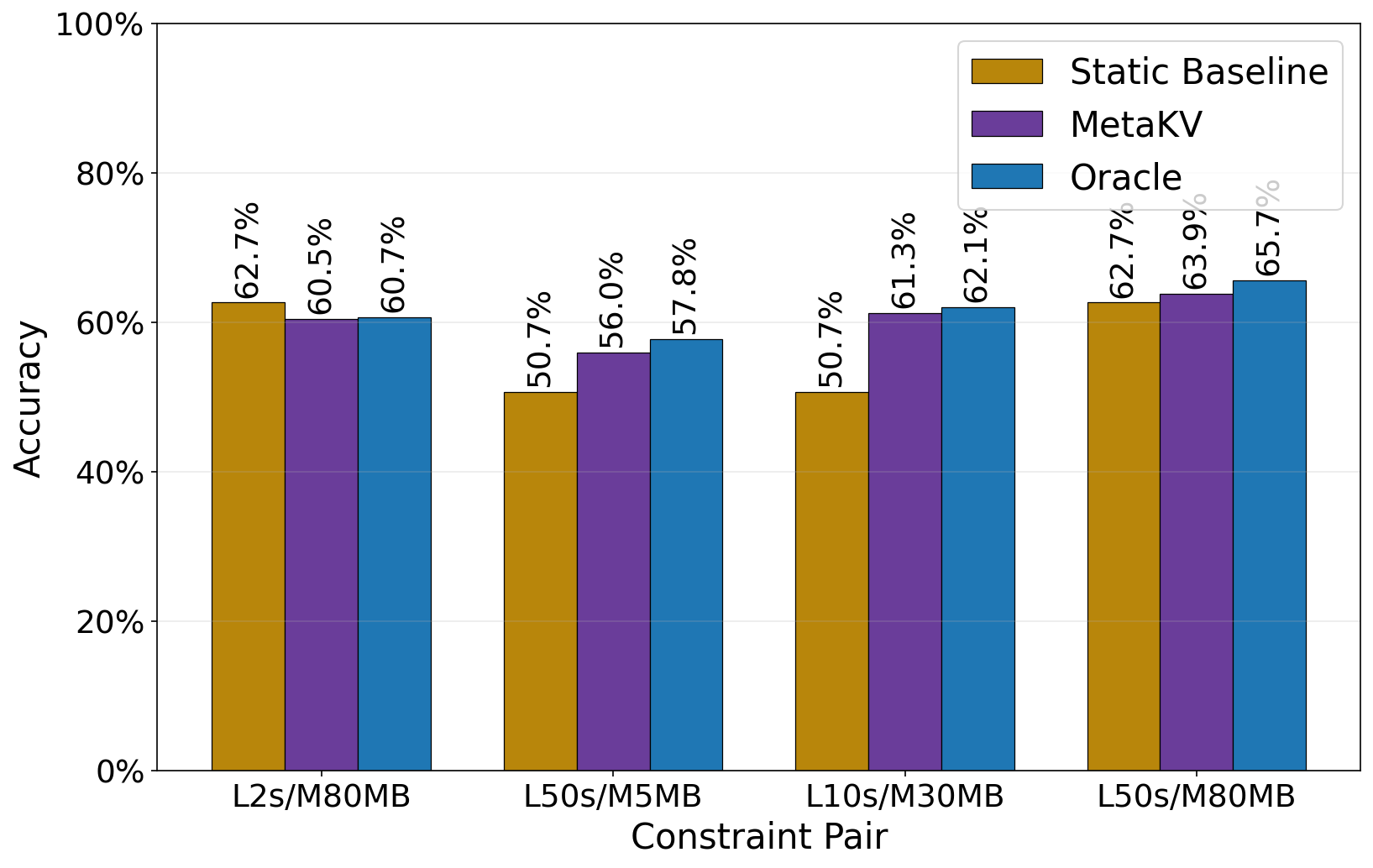}

        \textbf{(c)}
    \end{subfigure}

    \caption{Comparison of \static, \metakv, and
            \oracle across four representative latency--memory constraint settings.
            (a) Constrained success rate (CSR), shown as a proportion. (b) Constraint violation rate, the percentage
            of prompts exceeding the latency or peak memory budget. (c) Accuracy, shown as a percentage.}
    \label{fig:corner-comparisons}
\end{figure*}

\noindent \paragraph{\metakv adapts to different resource bottlenecks.}
To understand why \metakv achieves higher CSR than the best static configuration, we compare accuracy and constraint violations under four representative constraint settings shown in \autoref{fig:corner-csr}: tight latency with loose peak memory ($L_{\max}=2\,\text{s}, M_{\max}=80\,\text{MB}$), loose latency with tight peak memory ($L_{\max}=50\,\text{s}, M_{\max}=5\,\text{MB}$), moderate latency and peak memory ($L_{\max}=10\,\text{s}, M_{\max}=30\,\text{MB}$), and loose latency and peak memory ($L_{\max}=50\,\text{s}, M_{\max}=80\,\text{MB}$).


Under tight latency and loose peak memory constraints, \metakv improves CSR from 0.314 to 0.366. As shown in \autoref{fig:corner-violations}, \metakv reduces the constraint violation rate from $45.9\%$ to $35.8\%$, while its accuracy is only $2.2$ percentage points lower than the static configuration (\autoref{fig:corner-accuracy}). The \static selects H$_2$O 60\% for all prompts, which violates the tight latency constraint more often. In contrast, \metakv can select lower-latency configurations for prompts where H$_2$O 60\% would exceed the budget.


Under tight peak memory and loose latency constraints, \metakv improves accuracy by $5.3$ percentage points over the \static, while increasing the constraint violation rate by only $3.1$ percentage points. The \static achieves fewer violations because it uses H$_2$O 20\% for every prompt, an aggressive compression setting that reduces memory usage but also lowers accuracy. But \metakv can select less aggressive and more accurate configurations for shorter prompts that still satisfy the peak memory budget.


Under moderate latency and peak memory constraints, \metakv improves accuracy by $10.6$ percentage points while increasing the constraint violation rate by only $0.9$ percentage points compared to the \static. The \static uses H$_2$O 20\% for all prompts, which keeps latency and peak memory low but sacrifices accuracy. But \metakv can select more accurate configurations for prompts that still satisfy the constraints, leading to a large accuracy gain with only a small increase in violations.


Under loose latency and peak memory constraints, all methods achieve a $0\%$ constraint violation rate because a feasible configuration exists for every prompt. In this setting, \metakv improves accuracy by $1.2$ percentage points over the \static. The \static uses H$_2$O 60\% for all prompts, while \metakv can select more accurate configurations for shorter prompts that have enough latency and peak memory headroom.


Although violation rates reach up to $38\%$ in some settings, most are not caused by \metakv's selection pipeline. As shown in \autoref{fig:corner-violations}, the \oracle also violates up to $35\%$ of prompts, which occurs only when none of the available configurations can satisfy the given constraints. \metakv's violation rate is only $0$--$6$ percentage points higher than the \oracle, indicating that most violations are due to limitations of the current compression method pool rather than prediction errors. We discuss how to mitigate this issue further in \autoref{sec:discussion}. Similarly, \metakv's accuracy is only $0.2$--$1.8$ percentage points lower than the \oracle, showing that its selection policy performs close to the best achievable accuracy among the available configurations.

\textit{\textbf{Key Takeaways:} \metakv improves CSR by adapting to the active resource bottleneck, reducing violations under tight constraints, and improving accuracy when more configurations are feasible.}

\noindent \paragraph{\metakv relies on multiple compression configurations.}

\begin{figure}[t]
    \centering
    \includegraphics[width=0.9\linewidth]{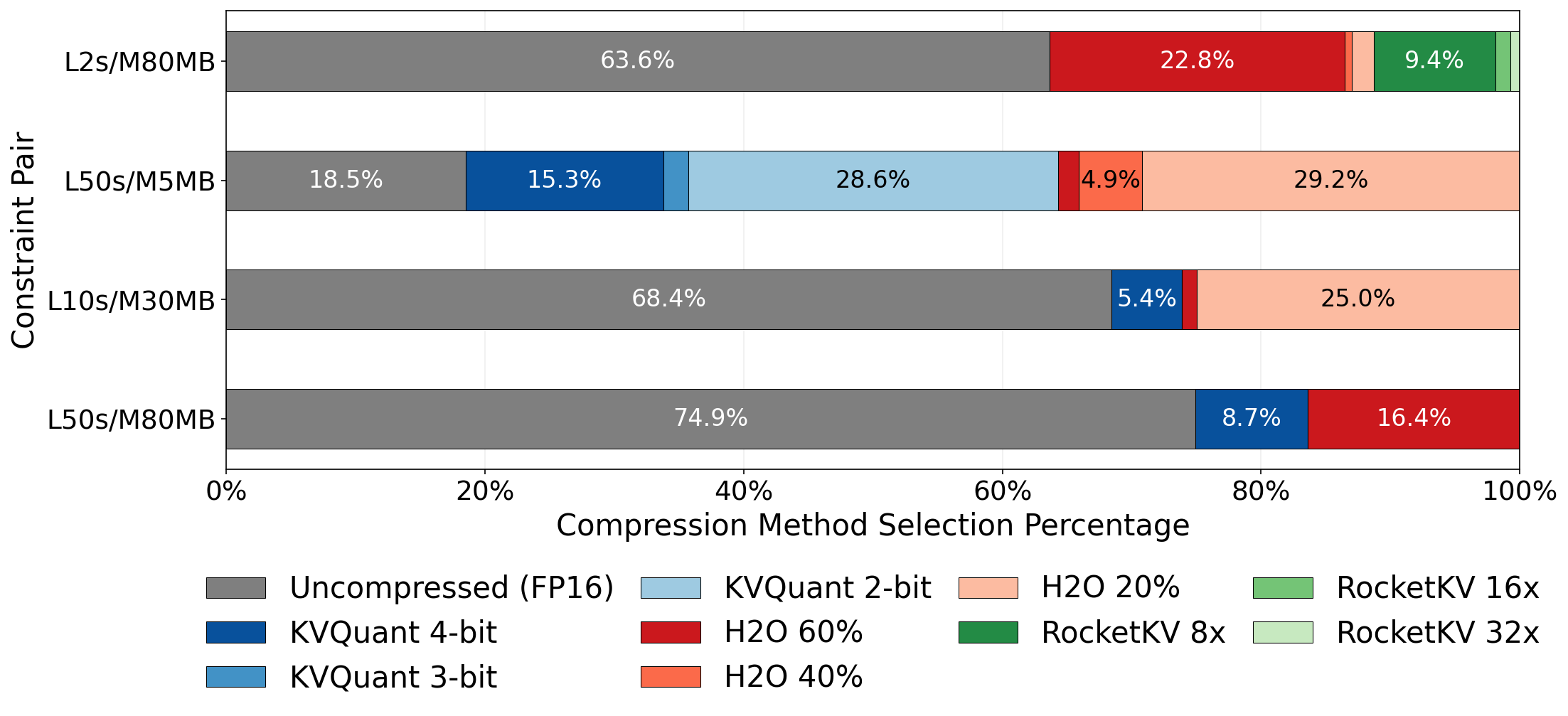}
    \caption{Proportion of prompts assigned to each configuration by \metakv
            under four representative latency--memory constraint pairings.}
    \label{fig:metakv-selection-proportions}
\end{figure}

To understand how \metakv adapts its selections across deployment settings, \autoref{fig:metakv-selection-proportions} shows the fraction of prompts assigned to each configuration under the four representative constraint pairs.

Under loose latency and peak memory constraints ($L_{\max}=50\,\text{s}, M_{\max}=80\,\text{MB}$), \metakv selects the uncompressed FP16 configuration for $74.9\%$ of prompts, prioritizing accuracy when compression is unnecessary. The remaining prompts use H$_2$O 60\% ($16.4\%$) or KVQuant 4-bit ($8.7\%$), mainly when longer contexts require mild compression to remain within the peak memory budget.

When latency is tight ($L_{\max} = 2\,\text{s}$, $M_{\max} = 80\,\text{MB}$), FP16 usage drops to 63.6\% and \metakv shifts toward low-overhead methods. Some prompts exceed the 80\,MB budget under FP16, so \metakv routes them to H$_2$O 60\%, taking up 22.8\% of selections. For other prompts, where FP16 would instead exceed the tight latency budget, \metakv leverages RocketKV's lowest latency on ARC-Challenge (shown in \autoref{app:additional-results}), selecting it 9.8\% of the time.

Under tight peak memory constraints ($L_{\max} = 50\,\text{s}, M_{\max} = 5\,\text{MB}$), FP16 selection drops to 18.5\% because it can only meet the peak memory threshold for the shortest prompts while using FP16. Instead, \metakv distributes selections across aggressive compression settings: H$_2$O 20\% (29.2\%), KVQuant 2-bit (28.6\%), and KVQuant 4-bit (15.3\%). This balanced split illustrates how \metakv chooses different methods based on a prompt's needs: it routes to KVQuant when prompt context requires preserving key tokens and to H$_2$O when it needs lower latency.

Finally, under moderate constraints ($L_{\max} = 10\,\text{s}, M_{\max} = 30\,\text{MB}$), \metakv returns to FP16 for 68.4\% of prompts, while relying on H$_2$O 20\% (25.0\%) and KVQuant 4-bit (5.4\%) to handle longer inputs without exceeding mid-tier budgets.

\textit{\textbf{Key Takeaways:} \metakv does not rely on a single dominant strategy; instead, it dynamically reconfigures its selection to FP16 under relaxed peak memory limits, fast sparse-attention pruning under strict latency, and dense quantization or token eviction under tight peak memory ceilings.}

\section{Discussion and Limitations}
\label{sec:discussion}

\noindent\textbf{Supporting new compression methods.}
\metakv is not tied to a fixed set of KV cache compression methods. New methods can be added to the candidate pool by profiling them offline and retraining the prediction models, without changing \metakv's core selection logic.

\noindent\textbf{Adapting to new models and hardware.}
\metakv's latency and peak memory predictors are specific to the deployment environment. Using another LLM or hardware platform requires re-profiling the candidate configurations and retraining the predictors. Reducing this recalibration cost through transfer learning or lightweight online calibration is an important direction for future work.

\noindent\textbf{Handling constraint violations.}
\metakv relies on predicted latency and peak memory, so some unseen prompts may still violate the given constraints. Our results show that most violations occur when none of the available configurations can satisfy the constraints, while a smaller fraction is caused by prediction error. In practice, violations can be reduced by expanding the candidate pool, using conservative prediction margins, relaxing the constraints, or provisioning additional resources.

\noindent\textbf{Supporting additional constraints.}
This work focuses on end-to-end latency and peak KV cache memory, but \metakv can be extended to other deployment metrics such as time-to-first-token, throughput, inference cost, or energy consumption. These metrics can be collected during offline profiling, predicted at runtime, and incorporated into the selection policy as additional constraints or objectives.

\noindent\textbf{Scope of the evaluation.}
Our evaluation covers one LLM, one hardware platform, four datasets, and three KV cache compression methods. Therefore, the results should not be interpreted as showing that any individual compression method is universally better than the others. Instead, they demonstrate that different configurations provide different accuracy, latency, and peak memory trade-offs, and that adapting the configuration at the prompt level can improve constrained performance. Evaluating \metakv across additional models, hardware platforms, workloads, and compression methods remains future work.

\clearpage
\bibliographystyle{plain}
\bibliography{reference}

\clearpage

\appendix
\section{Per-Dataset Results}
\label{app:additional-results}

\begin{table*}[h]
  \centering
  \caption{Accuracy, average end-to-end latency, and average peak KV-cache
  memory for each configuration on the held-out test split of each dataset
  (308 prompts per dataset).}
  \label{tab:per-dataset-results}
  \resizebox{\textwidth}{!}{%
  \begin{tabular}{llccc}
    \toprule
    Dataset & Configuration & Accuracy & Latency (s) & Peak Memory (MB) \\
    \midrule

    \multirow{10}{*}{GSM8K}
      & Uncompressed   & 0.6461 & 1.1977 & 95.9923 \\
      & KVQuant 2-bit       & 0.3896 & 14.0340 & 17.3521 \\
      & KVQuant 3-bit       & 0.5909 & 14.9944 & 23.9535 \\
      & KVQuant 4-bit       & 0.6136 & 15.5878 & 30.8855 \\
      & H$_2$O 20\%         & 0.5325 & 1.6776 & 17.2394 \\
      & H$_2$O 40\%         & 0.6234 & 1.7112 & 34.5990 \\
      & H$_2$O 60\%         & 0.6364 & 1.6912 & 51.9269 \\
      & RocketKV $8\times$  & 0.6201 & 5.2331 & 87.2135 \\
      & RocketKV $16\times$ & 0.5065 & 4.8067 & 86.7212 \\
      & RocketKV $32\times$ & 0.3409 & 5.8915 & 86.7212 \\
    \midrule

    \multirow{10}{*}{ARC-Challenge}
      & Uncompressed   & 0.5617 & 2.1265 & 4.5763 \\
      & KVQuant 2-bit       & 0.5260 & 19.5933 & 1.1000 \\
      & KVQuant 3-bit       & 0.5260 & 20.8838 & 1.6426 \\
      & KVQuant 4-bit       & 0.5552 & 21.8262 & 2.4448 \\
      & H$_2$O 20\%         & 0.3377 & 2.8306 & 0.8287 \\
      & H$_2$O 40\%         & 0.5130 & 2.6755 & 1.7792 \\
      & H$_2$O 60\%         & 0.5552 & 2.5428 & 2.7013 \\
      & RocketKV $8\times$  & 0.3994 & 1.3618 & 3.8121 \\
      & RocketKV $16\times$ & 0.3084 & 1.3924 & 3.8121 \\
      & RocketKV $32\times$ & 0.3247 & 1.4142 & 3.8121 \\
    \midrule

    \multirow{10}{*}{HellaSwag}
      & Uncompressed   & 0.7987 & 4.6149 & 10.5698 \\
      & KVQuant 2-bit       & 0.7727 & 42.9805 & 2.1627 \\
      & KVQuant 3-bit       & 0.7955 & 44.8244 & 3.0977 \\
      & KVQuant 4-bit       & 0.7987 & 46.9775 & 4.2995 \\
      & H$_2$O 20\%         & 0.7208 & 6.1472 & 2.0341 \\
      & H$_2$O 40\%         & 0.7857 & 5.7964 & 4.1778 \\
      & H$_2$O 60\%         & 0.7955 & 5.5188 & 6.3109 \\
      & RocketKV $8\times$  & 0.4286 & 7.2581 & 6.1530 \\
      & RocketKV $16\times$ & 0.2792 & 7.4863 & 6.1530 \\
      & RocketKV $32\times$ & 0.2987 & 7.4716 & 6.1530 \\
    \midrule

    \multirow{10}{*}{SQuAD 1.1}
      & Uncompressed   & 0.5747 & 0.1139 & 26.7459 \\
      & KVQuant 2-bit       & 0.4545 & 2.1050 & 4.9320 \\
      & KVQuant 3-bit       & 0.5227 & 2.2765 & 6.9417 \\
      & KVQuant 4-bit       & 0.5519 & 2.2714 & 9.2091 \\
      & H$_2$O 20\%         & 0.4383 & 0.1362 & 5.0657 \\
      & H$_2$O 40\%         & 0.4870 & 0.1451 & 10.2557 \\
      & H$_2$O 60\%         & 0.5195 & 0.1484 & 15.4213 \\
      & RocketKV $8\times$  & 0.4058 & 0.3886 & 25.8920 \\
      & RocketKV $16\times$ & 0.3149 & 0.4384 & 25.9054 \\
      & RocketKV $32\times$ & 0.1688 & 0.5426 & 25.8892 \\

    \bottomrule
  \end{tabular}%
  }
\end{table*}

\end{document}